\documentclass[lettersize,journal]{IEEEtran}
\usepackage{amsmath,amsfonts}
\usepackage{algorithmic}
\usepackage{algorithm}
\usepackage{array}
\usepackage[caption=false,font=normalsize,labelfont=sf,textfont=sf]{subfig}
\usepackage{textcomp}
\usepackage{stfloats}
\usepackage{url}
\usepackage{verbatim}
\usepackage{graphicx}
\usepackage{cite}
\usepackage{xcolor}
\usepackage{booktabs}  
\usepackage{multirow}  
\usepackage{lipsum}
\usepackage{bbding}
\usepackage{bm} 
\usepackage{mathrsfs} 
\usepackage{graphicx}
\usepackage{xcolor}
\usepackage{booktabs}
\usepackage{rotating}
\usepackage{amssymb}
\usepackage{algorithm}    
\usepackage{algorithmic}  

\usepackage[colorlinks,
linkcolor=red,
anchorcolor=blue,
citecolor=green
]{hyperref}

\begin{document}
\title{When the Past Matters: FlashBack Memory for Precipitation Nowcasting}
	
\author{Yuhao Du, Boxiao Huang, Chengrong Wu, Jiankai Zhang$^{*}$ \thanks{* Corresponding author}
\thanks{This research was jointly supported by the National Natural Science Foundation of China (U2442211). The computations in this research was supported by Supercomputing Center of Lanzhou University. The source code will be released at \href{https://github.com/duyhlzu/Flash-Back-Memory}{https://github.com/duyhlzu/Flash-Back-Memory}.}
\thanks{Yuhao Du and Jiankai Zhang are with the College of Atmospheric Sciences, Lanzhou University, Lanzhou 730000, China (e-mail: duyh2024@lzu.edu.cn; wying2024@lzu.edu.cn; jkzhang@lzu.edu.cn).}
\thanks{Boxiao Huang is with the Fuqua School of Business, Duke University, North Carolina 27705, USA (e-mail: boxiao.huang@duke.edu).}
\thanks{Chengrong Wu is with the Department of Computer Science, University of Manchester, Manchester M139PL, UK (e-mail: chengrong.wu@student.Manchester.ac.uk).}
	}

\markboth{Journal of \LaTeX\ Class Files,~Vol.~14, No.~8, August~2021}%
{Shell \MakeLowercase{\textit{et al.}}: A Sample Article Using IEEEtran.cls for IEEE Journals}


\maketitle

\begin{abstract}
Accurate precipitation nowcasting is crucial for disaster mitigation and socio-economic planning, yet existing methods often struggle with false alarms, missed events, and long-range dependency modeling at high spatiotemporal resolution. To address these challenges, we propose FlashBack Memory (FB), a module that dynamically retrieves key historical states and integrates them via an adaptive fusion gate, enhancing the spatiotemporal representation capability of recurrent-based models. We incorporate FB into PredRNN, PredRNNpp, MIM, MotionRNN, and PredRNN-V2, and evaluate on CIKM2017, Shanghai2020, and SEVIR datasets. Experimental results demonstrate that FB significantly improves MSE, MAE, SSIM, and CSI metrics, particularly for high-intensity rainfall and long-sequence predictions, while reducing false alarms and missed events and enhancing temporal consistency and spatial localization. The proposed method provides a general and efficient memory enhancement mechanism, improving the overall performance of recurrent-based precipitation nowcasting models.
\end{abstract}

\begin{IEEEkeywords}
Precipitation Nowcasting, Recurrent Neural Networks (RNNs), Dynamic Memory Retrieval, Spatiotemporal Modeling
\end{IEEEkeywords}

\section{Introduction}

\IEEEPARstart{A}{ccurate} and timely precipitation nowcasting plays a crucial role in safeguarding lives and property as well as optimizing socioeconomic activities\cite{b1,b2,b3,b4}. It underpins the establishment of severe weather early warning systems, enhances disaster response decision-making, and improves operational efficiency in sectors such as transportation, agriculture, and energy management\cite{b4,b5,b6}. With the increasing frequency of extreme weather events driven by climate change, the demand for reliable short-term precipitation forecasts with high spatiotemporal resolution has become more urgent than ever\cite{b6,b7}.

Traditional precipitation nowcasting methods, such as Numerical Weather Prediction (NWP)\cite{b8}, have achieved remarkable progress but still exhibit significant limitations at fine spatiotemporal scales. The inherent complexity of atmospheric processes, sensitivity to initial conditions, and high computational costs make it difficult for NWP to accurately depict rapidly evolving phenomena such as convective precipitation, thereby constraining its ability to deliver high-quality short-term forecasts in real-time applications.

To overcome these shortcomings, data-driven approaches—especially deep learning (DL) methods—have emerged as powerful alternatives. Current mainstream forecasting models can be broadly categorized into two types: deterministic models\cite{b9,b10} and deep generative models\cite{b11,b12}. The former directly learn the mapping between inputs and outputs to provide deterministic predictions, making them suitable for efficient, real-time short-term forecasting, whereas the latter model conditional probability distributions to generate diverse possible outcomes, offering better representation of uncertainty and complex distributions. This study primarily focuses on improving deterministic precipitation nowcasting models, aiming to enhance their ability to capture long-range spatiotemporal dependencies and complex weather evolutions while maintaining real-time performance and accuracy.

In the task of precipitation nowcasting, models are required not only to capture short-term local variations but also to understand the long-term evolution of weather patterns. When analyzing sequences of radar images, human meteorological experts do not make predictions based solely on the most recent frames. Instead, they flexibly recall relevant historical patterns that resemble the current phenomena, allowing them to infer future developments. This cognitive process is \textbf{selective} and \textbf{context-aware}: experts do not review all past information uniformly but instinctively retrieve the most relevant memories.

Although recent advances in RNN-based models (such as ConvLSTM\cite{b9}, PredRNN\cite{b10}, MotionRNN\cite{b13}, SwinLSTM\cite{b14} and VMRNN\cite{b15}) have led to remarkable progress in precipitation forecasting, these models still encounter challenges in long-term spatiotemporal modeling. Specifically, due to intrinsic limitations such as \textbf{catastrophic forgetting} and constrained ability to model long-range dependencies, the performance of RNN-based models tends to plateau when tackling longer forecasting horizons. Moreover, existing models typically adopt fixed schemes for aggregating historical states, lacking a flexible memory retrieval mechanism that dynamically associates past information with the current input. As a result, they often fail to retain critical information in scenarios involving complex weather dynamics.

To address these issues and better emulate the retrospective reasoning process employed by human experts, we propose \textbf{FlashBack Memory}---a memory module that dynamically retrieves key historical states conditioned on the current input. The design of FlashBack Memory incorporates the following three core innovations:

\begin{itemize}
	\item \textbf{Dynamic Memory Retrieval}: FlashBack Memory adaptively selects the most relevant past states based on the current input, rather than indiscriminately stacking all historical information.
	
	\item \textbf{Dynamic Routing Aggregation}: Inspired by the idea of dynamic routing, we introduce a lightweight iterative optimization mechanism that gradually enhances the model’s focus on the most valuable memory fragments.
	
	\item \textbf{Adaptive Fusion Gate}: To effectively integrate the retrieved memories with the current hidden state, we design an adaptive fusion gate, which allows the model to dynamically balance the contributions of both components, yielding a more refined and context-aware hidden representation.
\end{itemize}

By introducing FlashBack Memory, the model not only acquires enhanced capabilities for capturing long-term dependencies but also gains the flexibility to selectively retrieve information most conducive to the current prediction task. In this study, we integrate FlashBack Memory into the Spatiotemporal-LSTM framework and conduct systematic evaluations on three representative precipitation nowcasting datasets---\textbf{CIKM2017}, \textbf{Shanghai2020}, and \textbf{SEVIR}. Experimental results demonstrate that our proposed method significantly improves forecasting accuracy, particularly over longer prediction horizons and under complex weather conditions.

\section{Related Work}

Current video- and radar-based precipitation nowcasting models can be broadly grouped into two families: recurrent-based methods and recurrent-free methods. Each family adopts different architectural priors and provides distinct remedies for the core challenge of modeling long-range spatiotemporal dependencies.

Recurrent-based spatiotemporal prediction methods are exemplified by ConvLSTM, which embeds temporal gating into convolutional units to simultaneously capture spatial and temporal features. Subsequent research has evolved along three main directions: mitigating long-range dependencies, enhancing global context modeling, and improving memory selectivity. In terms of memory mechanisms, some studies introduce interleaved or multi-stream spatiotemporal memory units to facilitate gradient propagation across time and extend effective memory\cite{b10,b16,b17,b18,b19,b20}. For state modeling, a series of methods employ explicit motion decomposition, separation of trend and transient components, dynamic gating, or weighted mechanisms to reduce the interference of short-term noise on long-term states and better capture complex motion evolution\cite{b13,b21}. Regarding global context modeling, researchers gradually incorporate multi-scale convolutions, sparse routing, local–global self-attention, or hierarchical retrieval into recurrent units\cite{b4,b7,b14,b15}, thereby enhancing the modeling of long-range spatiotemporal dependencies while preserving the computational efficiency of recurrence.

In contrast, recurrent-free models discard stepwise state propagation and directly perform end-to-end modeling on spatiotemporal tensors. Early works such as SimVP\cite{b22} employ deep convolutions and multi-scale stacking to transform precipitation nowcasting into a purely feedforward mapping task, significantly improving parallelization efficiency. Subsequent research has further advanced these models along aspects such as long-range dependency modeling, global context capture, and dynamic information selection. Some approaches introduce cross-time convolutions or temporal attention modules to explicitly model long-distance spatiotemporal dependencies\cite{b23}, alleviating the gradient decay problem in long-sequence prediction characteristic of recurrent methods. Others leverage multi-view decomposition or multi-modal fusion to separate local dynamics from global trends, thereby enhancing the representation of complex weather evolution\cite{b24,b25}. The latest PredFormer\cite{b26} series combines spatiotemporal bidirectional attention with dynamic token selection in a Transformer framework, further strengthening the capture of critical patterns and long-range dependencies while maintaining high parallel efficiency.

Although both recurrent and recurrent-free models have achieved significant progress in spatiotemporal sequence modeling, they still face inherent limitations in long-range dependency modeling and memory retrieval. Recurrent methods, despite incorporating multi-stream memories, explicit motion decomposition, or local–global attention to extend effective memory, still rely on fixed sequential propagation, making them prone to gradient decay and key state forgetting. Recurrent-free models, on the other hand, achieve parallelization and a global receptive field through cross-time convolutions, attention, and multi-modal decomposition but generally lack explicit dynamic memory modules, resulting in historical information being utilized more as “average aggregation” rather than “selective retrieval.” To address this shared limitation, we propose FlashBack Memory, motivated by the retrospective reasoning process employed by meteorological experts. During prediction, it dynamically retrieves historical states most relevant to the current input, and integrates these key memories via dynamic routing and an adaptive fusion gate. This design allows flexible, context-aware long-range dependency modeling while maintaining efficient prediction. Compared with existing methods, FlashBack Memory moves beyond relying solely on sequential hidden state propagation or uniform aggregation, providing a conditionally driven explicit historical selection mechanism that preserves and leverages critical temporal memories in complex weather evolution scenarios, thereby significantly improving the modeling of long-range dependencies and non-stationary dynamics.

\section{Method}
\subsection{Problem Formulation}

We address the task of precipitation nowcasting, where the objective is to forecast upcoming radar reflectivity maps based on a sequence of historical observations. Let $\mathcal{I}_{1:T} \in \mathbb{R}^{C \times T \times H \times W}$ denote the observed input sequence of $T$ frames, where $C$ is the number of channels, and $H \times W$ is the spatial resolution. The goal is to predict the future sequence $\widehat{\mathcal{I}}_{T+1:T+T'} \in \mathbb{R}^{C \times T' \times H \times W}$ over a forecasting horizon of $T'$ steps.

To this end, we aim to learn a predictive function $\mathcal{F}_\Theta: \mathbb{R}^{C \times T \times H \times W} \rightarrow \mathbb{R}^{C \times T' \times H \times W}$, parameterized by $\Theta$, such that:

\begin{equation}
	\widehat{\mathcal{I}}_{T+1:T+T'} = \mathcal{F}_\Theta(\mathcal{I}_{1:T}),
\end{equation}

where $\widehat{\mathcal{I}}$ denotes the predicted future radar frames. The model parameters $\Theta$ are optimized by minimizing a discrepancy between predictions and the corresponding ground truth sequence $\mathcal{I}_{T+1:T+T'}$ using a task-specific loss function $\mathcal{L}$:

\begin{equation}
	\Theta^\star = \arg\min_{\Theta} \mathcal{L}\left( \mathcal{F}_\Theta(\mathcal{I}_{1:T}), \mathcal{I}_{T+1:T+T'} \right).
\end{equation}

Unlike conventional video prediction tasks, precipitation nowcasting poses unique challenges due to its highly dynamic, multi-scale spatiotemporal dependencies and the necessity for long-term reasoning. Human forecasters often rely on prior experiences and analogical recall, revisiting previously seen patterns that resemble current developments. Inspired by this intuition, we extend the standard formulation by incorporating a recurrent memory mechanism that dynamically retrieves and integrates relevant historical states, beyond the fixed input window $\mathcal{I}_{1:T}$.

In this work, we introduce a memory-augmented prediction model that enriches the predictive process with selective flashback retrieval. This allows the model to maintain a flexible, context-aware temporal scope, enabling it to reference and integrate critical past information as needed, thereby enhancing its long-term forecasting capability.

\subsection{Dynamic Memory Retrieval}

\begin{figure}[htbp]
	\centerline{\includegraphics[width=8cm]{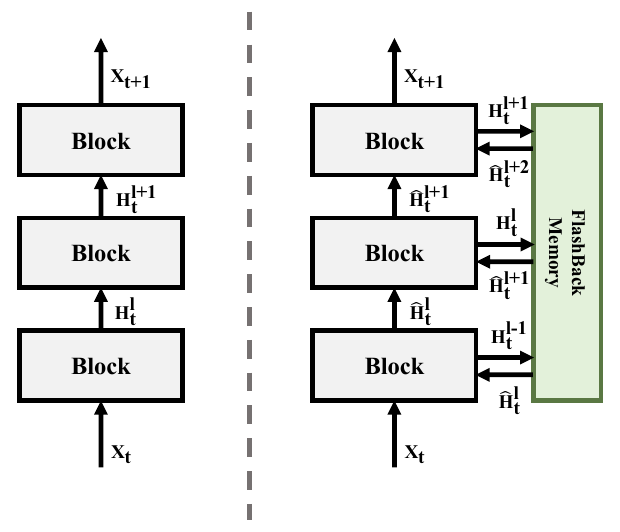}}
	\caption{Structural Comparison Between ConvRNN and ConvRNN+FB.
	}
	\label{fig:one}
\end{figure}

To address the degradation of long-range dependency modeling in recurrent structures, we introduce a content-aware memory retrieval module that allows selective access to historical hidden states. Unlike conventional RNNs that compress all temporal information into a single hidden vector, our design explicitly maintains a dynamic external memory $\mathcal{M}$, enabling flexible retrieval and aggregation of relevant past states. This memory mechanism is particularly beneficial in spatiotemporal scenarios such as video prediction, where multiple past observations may share visual similarities but differ significantly in semantic relevance.We present a comparison between the standard recurrent neural network framework and the FlashBack-based network framework in Figure \ref{fig:one}.

Let $\mathbf{h} \in \mathbb{R}^{B \times C \times H \times W}$ be the current hidden state at the decoding step, where $B$ is the batch size, $C$ is the number of hidden channels, and $H \times W$ are the spatial dimensions. During the encoding phase, we maintain a dynamic memory buffer:
\begin{equation}
	\mathcal{M} = \{ \mathbf{h}_1, \mathbf{h}_2, \dots, \mathbf{h}_K \}, \quad \mathbf{h}_k \in \mathbb{R}^{B \times C \times H \times W}.
\end{equation}
where $K$ denotes the number of memory slots (up to a maximum of $K_{\text{max}} = 15$ in our implementation). Each stored memory is detached from gradient flow to avoid backpropagation through the full history, matching the implementation:
\texttt{self.memory.append(h.detach())}.

To retrieve relevant memories, we first flatten the current state and memory entries into a vector space for computing cosine similarity. We use the following reshape operation for both the current state and each memory tensor:
\begin{align}
	\mathbf{h}_k^{flat} &= \text{Flatten} \left(\mathbf{h}_k \right),
\end{align}
where $\text{Flatten}(\mathbf{h})$ refers to the operation that reshapes the tensor $\mathbf{h} \in \mathbb{R}^{B \times C \times H \times W}$ into a 2D matrix of size $\mathbb{R}^{B \times D}$, with $D = C \cdot H \cdot W$ representing the total number of spatial features.

Similarly, we define the flatten memory buffer \( \mathcal{M}^{flat} \in \mathbb{R}^{B \times K \times D}\) as:

\begin{equation}
	\mathcal{M}^{flat} = \{ \mathbf{h}^{flat}_1, \mathbf{h}^{flat}_2, \dots, \mathbf{h}^{flat}_K \}, \quad \mathbf{h}^{flat}_k \in \mathbb{R}^{B \times D}.
\end{equation}

Then, the cosine similarity between the current hidden state and each memory slot is computed across the batch dimension. Specifically, for each $k \in \{1, \dots, K\}$:
\begin{equation}
	s_k^{(b)} = \cos\left( \mathbf{h}_{K+1}^{flat, (b)},\ \mathbf{h}_k^{flat, (b)} \right), \quad \forall b \in \{1, \dots, B\},
\end{equation}
where, \( s_k^{(b)} \in \mathbb{R}^{B \times D} \) refers to similarity scores, \( \mathbf{h}_{K+1} \) refers to the input hidden state in this moment.

We can aggregate these similarities into a matrix:
\begin{equation}
	\mathbf{S} = \left[ s_k^{(b)} \right] ,\quad \mathbf{S} \in \mathbb{R}^{B \times K \times D}.
\end{equation}

Given two flattened hidden states $\mathbf{h}_a, \mathbf{h}_b \in \mathbb{R}^{1 \times D}$; $ a,b \in \{1, \dots, K\}$, their normalized cosine similarity is computed as:
\begin{equation}
	\cos(\mathbf{h}_a, \mathbf{h}_b) = \frac{\langle \mathbf{h}_a, \mathbf{h}_b \rangle}{\|\mathbf{h}_a\|_2 \cdot \|\mathbf{h}_b\|_2}
	= \left\langle \frac{\mathbf{h}_a}{\|\mathbf{h}_a\|_2}, \frac{\mathbf{h}_b}{\|\mathbf{h}_b\|_2} \right\rangle.
\end{equation}

The similarity scores $\{ s_k^{(b)} \}_{k=1}^K$ for each sample $b$ are converted into normalized attention weights via softmax:
\label{eq:a}
\begin{equation}
	\boldsymbol{\alpha}_k^{(b)} = \frac{\exp\left(s_k^{(b)}\right)}{\sum_{j=1}^K \exp\left(s_j^{(b)}\right)}, \quad \boldsymbol{\alpha}^{(b)} \in \mathbb{R}^{B \times K}.
\end{equation}

This operation yields a per-sample, per-slot attention vector $\boldsymbol{\alpha}^{(b)}$, guiding how much each memory contributes to the final recall.

Finally, the attended memory representation for each batch $b$ is computed as the weighted sum over all historical memory slots:
\label{eq:b}
\begin{equation}
	\mathbf{h}_{\text{mem}}^{(b)} = \sum_{k=1}^{K} Enc(\alpha_k^{(b)}) \cdot \mathbf{h}_k^{(b)},
\end{equation}
where, \( Enc \)$(\cdot)$ refers to expanding a tensor of size $\mathbb{R}^{B \times K}$ to $\mathbb{R}^{B \times K \times 1 \times 1 \times 1}$, \( \mathbf{h}_{\text{mem}} \in \mathbb{R}^{B \times C \times H \times W} \) are refers to the \textbf{FlashBack Memory}. This retrieval is differentiable and dynamically adapts to the current context, enabling the decoder to incorporate long-range dependencies from the most relevant prior observations.

\subsection{Dynamic Routing Aggregation}

To further enhance the selectivity and robustness of memory retrieval, we augment the basic cosine attention mechanism with an iterative \textbf{dynamic routing} algorithm, inspired by capsule networks\cite{b27}. This mechanism enables progressive refinement of attention weights over multiple routing steps, allowing the model to focus more precisely on semantically relevant memory slots.

In conventional one-step soft attention, especially for spatiotemporal video data, multiple past states may appear visually similar but differ in semantic importance. Such ambiguity makes it difficult to assign accurate relevance weights. Dynamic routing mitigates this by iteratively adjusting attention based on the agreement between the current query and the retrieved memory summary.

Let $\mathbf{l} \in \mathbb{R}^{B \times K}$ be the initial similarity scores, initialized by the cosine similarities between the flattened current hidden state and each memory slot:
\begin{equation}
	\mathbf{l} = \cos\left( \mathbf{h}_{K+1}^{flat},\ \mathcal{M}^{flat} \right), \quad \mathbf{l} \in \mathbb{R}^{B \times K}
\end{equation}

For $R$ routing iterations (we use $R_{max}=3$), the attention weights $\boldsymbol{\alpha}$ are updated via softmax and used to compute a weighted sum of memory states:
\begin{equation}
	\alpha_k^{(r)} = \frac{\exp(\mathbf{l})}{\sum_{j=1}^K \exp(\mathbf{l}_j)},\quad \forall r \in \{1, \dots, R\}, \quad \boldsymbol{\alpha}^{(r)} \in \mathbb{R}^{B \times K}
\end{equation}

\begin{equation}
	\mathbf{h}_{\text{rout}}^{(r)} = \sum_{k=1}^K Enc( \alpha_k^{(r)} ) \cdot \mathbf{h}_k
\end{equation}

Then, we update the routing logits $\mathbf{t} \in \mathbb{R}^{B \times K}$ based on the agreement between the retrieved memory and each memory slot:
\begin{equation}
	\mathbf{t}_k \leftarrow \mathbf{t}_k +  Dec(\mathbf{h}_{\text{rout}}^{(R_{max})}) \cdot \mathcal{M}^{flat},
\end{equation}
where, \( Dec(\cdot) \) refers to reshape a tensor from \( \mathbb{R}^{B \times C \times H \times W} \) to \( \mathbb{R}^{B \times 1 \times D} \).
This process is repeated for $R$ iterations, gradually sharpening the focus on relevant memory slots through accumulated evidence.

Similar to Eq.(10), compute the routing Flashback Memory $\mathbf{h}^{rout}_{\text{mem}} \in \mathbb{R}^{B \times C \times H \times W}$:

\begin{equation}
	\mathbf{h}^{rout}_{\text{mem}} = \sum_{k=1}^{K} Enc(\boldsymbol{\beta}_k) \cdot \mathbf{h}_k,
\end{equation}
where,
\begin{equation}
	\boldsymbol{\beta}_k = \frac{\exp\left(\mathbf{t}_k\right)}{\sum_{j=1}^K \exp\left(\mathbf{t}_j\right)}, \quad \boldsymbol{\beta} \in \mathbb{R}^{B \times K}.
\end{equation}

\subsection{Adaptive Fusion Gate}
After $R$ iterations, we obtain the routing Flashback Memory $\mathbf{h}^{rout}_{\text{mem}}$ and proceed to fuse it with the current state using a gated mechanism:
\begin{equation}
	\mathbf{z} = \text{Concat}\left([ \mathbf{h}^{rout}_{\text{mem}}, \mathbf{h}_{K+1}^{flat} ]\right),
\end{equation}
\begin{equation}
	\mathcal{G} = \sigma\left( \mathbf{W}_{1 \times 1} * \mathbf{z} \right),
\end{equation}
where,   $\mathbf{z} \in \mathbb{R}^{B \times 2C \times H \times W}$, $\mathcal{G} \in \mathbb{R}^{B \times C \times H \times W}$, $*$ denote the convolution, $ \sigma $ donates the sigmoid function.
\begin{equation}
	\hat{\mathbf{h}} = \mathcal{G} \odot \mathbf{h}^{rout}_{\text{mem}} + (1 - \mathcal{G}) \odot \mathbf{h}_{K+1}^{flat}.
\end{equation}

Here, the convolutional gate $\mathcal{G}$ controls how much information to integrate from memory versus the current hidden state. The fused output $\hat{\mathbf{h}}$ is then used as the updated hidden state in the decoder.

The entire routing procedure is differentiable and can be trained end-to-end using stochastic gradient descent. Unlike traditional memory networks that rely on hard addressing or fixed attention, our method enables fine-grained, dynamic, and content-aware memory selection.

\subsection{Overall Architecture}

\begin{figure*}[htbp]
	\centerline{\includegraphics[width=18cm]{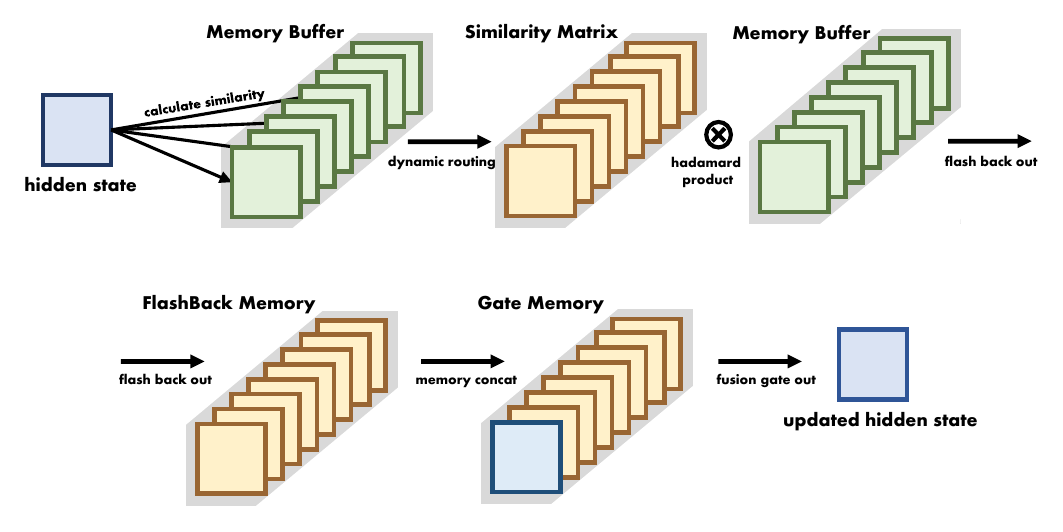}}
	\caption{Flowchart of the FlashBack Memory Module.
	}
	\label{fig:two}
\end{figure*}

To clearly illustrate the proposed \textbf{FlashBack} framework, we present its full inference procedure in Algorithm~\ref{alg:flashback} and depict the modular architecture in Figure~\ref{fig:two}. The method explicitly maintains historical hidden states in an external dynamic memory buffer $\mathcal{M}$ during the encoding phase and detaches them from gradient flow to reduce memory overhead. At decoding or prediction time, the current hidden state is first flattened and compared with each memory slot using normalized cosine similarity to obtain an initial retrieval via soft attention. Crucially, the multi-step dynamic routing iterations are applied \emph{before} Equation~(10) to progressively refine the attention weights, producing a high-quality aggregated memory feature $\mathbf{h}^{rout}_{\text{mem}}$. This feature is then fused with the current state through a gated mechanism to form the enhanced representation.

\begin{algorithm}[t]
	\caption{FlashBack Memory with Dynamic Routing}
	\label{alg:flashback}
	\small
	\begin{algorithmic}[1]
		\STATE \textbf{Input:} Current state $\mathbf{h}_{K+1} \in \mathbb{R}^{B \times C \times H \times W}$, 
		memory buffer $\mathcal{M} = \{\mathbf{h}_1,\dots,\mathbf{h}_K\}$, routing steps $R$
		\STATE \textbf{Output:} Refined state $\hat{\mathbf{h}}$
		
		\STATE Flatten current state: $\mathbf{h}_{K+1}^{flat} \gets \text{Flatten}(\mathbf{h}_{K+1})$
		\STATE Flatten memory: $\mathcal{M}^{flat} \gets \text{Flatten}(\mathcal{M})$
		\STATE Compute initial logits: $l_k \gets \cos(\mathbf{h}_{K+1}^{flat}, \mathbf{h}_k^{flat}) \quad \forall k$
		
		\FOR{$r = 1$ to $R$}
		\STATE Compute attention weights: $\alpha_k^{(r)} \gets \frac{\exp(l_k)}{\sum_{j=1}^{K} \exp(l_j)}$
		\STATE Aggregate memory: $\mathbf{h}_{\text{rout}}^{(r)} \gets \sum_{k=1}^{K} \alpha_k^{(r)} \mathbf{h}_k$
		\STATE Update logits by agreement: $l_k \gets l_k + \langle Dec(\mathbf{h}_{\text{rout}}^{(r)}), \mathbf{h}_k^{flat} \rangle$
		\ENDFOR
		
		\STATE Compute final routing weights: $\beta_k \gets \frac{\exp(l_k)}{\sum_{j=1}^{K} \exp(l_j)}$
		\STATE Compute routed memory: $\mathbf{h}_{\text{mem}}^{rout} \gets \sum_{k=1}^{K} \beta_k \mathbf{h}_k$
		\STATE Fuse with current state:
		\STATE \quad $\mathbf{z} \gets \text{Concat}[\mathbf{h}_{\text{mem}}^{rout}, \mathbf{h}_{K+1}]$
		\STATE \quad $\mathcal{G} \gets \sigma(\mathbf{W}_{1\times1} * \mathbf{z})$
		\STATE \quad $\hat{\mathbf{h}} \gets \mathcal{G} \odot \mathbf{h}_{\text{mem}}^{rout} + (1-\mathcal{G}) \odot \mathbf{h}_{K+1}$
		
		\STATE \textbf{Return:} $\hat{\mathbf{h}}$
	\end{algorithmic}
\end{algorithm}

This design alleviates the limitation of conventional recurrent networks that compress all temporal information into a single vector and enables selective recall of semantically relevant past states, leading to improved performance in spatiotemporal modeling tasks. Figure~\ref{fig:two} illustrates the interaction among modules, while Algorithm~\ref{alg:flashback} details the complete FlashBack process, including memory update, similarity computation, \textbf{dynamic routing}, and gated fusion.

\section{EXPERIMENTS}
\subsection{Datasets and Metrics}

This paper selects precipitation datasets with different spatiotemporal resolutions for experiments, including \textbf{Shanghai2020}\cite{b28}, \textbf{CIKM2017}\cite{b29}, and \textbf{SEVIR}\cite{b30}.

The Shanghai2020 dataset, released by the Shanghai Central Meteorological Observatory (SCMO) in 2020, records historical precipitation events in the Yangtze River Delta region over multiple years. Each sample consists of 20 consecutive radar echo frames spanning 3 hours; the first 10 frames are recorded at 6-minute intervals (as input) and the last 10 frames at 12-minute intervals (as prediction). We performed data preprocessing to remove overly sparse samples, and finally retained 20,133 samples, of which 15,000 are used for training and 5,133 for testing.

The CIKM2017 dataset, provided by the Shenzhen Meteorological Bureau as a competition dataset, contains desensitized radar data, including radar images at four altitude levels. Each image has a resolution of 101×101 square kilometers. The data are recorded at 6-minute intervals, covering 15 time steps (1.5 hours). In this study, we select 10,000 precipitation samples at the 0.5 km altitude level. For each sample, the first 5 frames are used as input and the subsequent 10 frames as prediction. All samples are split into training and validation sets with a 9:1 ratio.

The Storm EVent ImageRy (SEVIR) dataset is a spatiotemporal Earth observation dataset composed of image sequences covering an area of 384 km × 384 km over a 4-hour period. We extracted 15,000 rainfall samples for experiments, with 12,000 used for training and 3,000 for testing. Each sample contains 25 radar echo frames recorded at 10-minute intervals (a total of 250 minutes), where the first 10 frames are used as input and the remaining 15 frames as prediction.

Detailed information about the datasets is provided in Table \ref{tab:dataset_config}. It is worth noting that, due to computational resource limitations, we use bilinear interpolation to downsample the image resolution of all datasets to 64×64.

To comprehensively evaluate the predictive performance of our model, we adopt four widely used quantitative metrics: Mean Squared Error (MSE), Mean Absolute Error (MAE), Structural Similarity Index (SSIM), and Critical Success Index (CSI). MSE and MAE measure the numerical discrepancies between predicted and ground-truth values. SSIM assesses the similarity between predicted and true images in terms of structure and texture; this is particularly important for precipitation radar data, as it reflects the model’s ability to preserve the spatial continuity, shape, and intensity distribution of rainfall patterns. CSI, commonly used in meteorological event evaluation, quantifies the trade-off between detection and false alarms under a given threshold, thereby indicating the reliability of event predictions.

\begin{table*}[htbp]
	\centering
    \normalsize
	\caption{Training configurations for different datasets.}
	\label{tab:dataset_config}
	\begin{tabular}{lccccc}
		\toprule
		\textbf{Datasets} & \textbf{Task} & \textbf{Epochs} & \textbf{Patch/Batch Size} & \textbf{Learning Rate} & \textbf{$K_\text{max}$} \\
		\midrule
		Shanghai2020 & 10$\rightarrow$10 & 200 & 4/16 & 0.0003 & 10 \\
		CIKM2017 & 5$\rightarrow$10 & 100 & 4/16 & 0.0003 & 10 \\
		SEVIR & 10$\rightarrow$15 & 100 & 4/16 & 0.0003 & 15 \\
		\bottomrule
	\end{tabular}
\end{table*}

\subsection{Quantitative Evaluation}

To verify the performance improvement brought by FlashBack Memory to ConvRNN models, we selected several classical models as baselines, including PredRNN, PredRNNpp, MIM, MotionRNN, and PredRNN-V2. In addition, we also compared a variety of recurrent-based and recurrent-free models. Tables \ref{tab:shanghai2020}, \ref{tab:cikm2017}, and \ref{tab:sevir} present these quantitative results.

Table \ref{tab:shanghai2020} reports the quantitative results on the Shanghai2020 dataset. We observe that both recurrent-based and recurrent-free models exhibit distinct performance patterns in terms of MSE, MAE, SSIM, and CSI. In general, conventional recurrent architectures excel at modeling temporal dependencies, whereas recurrent-free models are more flexible in spatial feature extraction. However, after integrating the proposed FlashBack (FB) module into the baseline models, all metrics show consistent improvements.
For instance, with PredRNN, the MSE decreases from 3.3879 to 3.3246 and MAE from 40.841 to 39.619, while SSIM increases from 0.9471 to 0.9489. On the stronger PredRNNpp backbone, the MSE further drops from 2.8878 to 2.7818, SSIM rises to 0.9562, and the CSI scores at different thresholds also improve noticeably. Similar trends are observed for MIM, MotionRNN, and PredRNN-V2, where FB helps reduce MSE and MAE while enhancing SSIM and CSI metrics.
These results clearly demonstrate the generality and effectiveness of the proposed FlashBack module across diverse architectures, especially in mitigating long-term dependency degradation and improving structural consistency (SSIM) as well as critical rainfall detection performance (CSI).

\renewcommand{\arraystretch}{1.2} 
\begin{table*}[htbp]
	\centering
	\caption{Quantitative results on Shanghai2020 datasets.}
	\begin{tabular}{c|l|cccccccc}
		\specialrule{1.5pt}{1pt}{1pt}
		\textbf{Type} & \textbf{Model} & \textbf{Params} & \textbf{FLOPS} & \textbf{MSE} $\downarrow$ & \textbf{MAE} $\downarrow$ & \textbf{SSIM} $\uparrow$ & \textbf{CSI30} $\uparrow$ & \textbf{CSI40} $\uparrow$ & \textbf{CSI50} $\uparrow$ \\
		\specialrule{1pt}{1pt}{1pt} 
		\midrule
		\multirow{9}{*}{\rotatebox{90}{\textbf{Recurrent-based}}} 
		& ConvLSTM (NeurIPS’2015) & 15.1M & 56.79G & 4.4513 & 47.655 & 0.9319 & 0.7574 & 0.7495 & 0.7135 \\
		& PredRNN (NeurIPS’2017) & 23.83M & 0.116T & 3.3879 & 40.841 & 0.9471 & 0.7896 & 0.7836 & 0.7458 \\
		& PredRNNpp (ICML’2018) & 38.6M & 0.172T & \underline{2.8878} & 37.828 & \underline{0.9553} & \underline{0.8097} & \underline{0.8033} & \underline{0.7656} \\
		& MIM (CVPR’2019) & 38.04M & 0.179T & 3.1073 & 37.860 & 0.9527 & 0.8031 & 0.7966 & 0.7576 \\
		& PyDNet (CVPR’2020) & 3.1M & 15.32G & 11.279 & 79.775 & 0.8415 & 0.6141 & 0.5911 & 0.5447 \\
		& MotionRNN (CVPR’2021) & 26.9M & 0.129T & 3.3499 & 43.076 & 0.9463 & 0.7899 & 0.7826 & 0.7441 \\
		& MAU (NeurIPS’2021) & 4.48M & 17.79G & 9.6037 & 74.541 & 0.8591 & 0.6384 & 0.6196 & 0.5753 \\
		& PredRNN-V2 (TPAMI’2022) & 23.9M & 0.117T & 3.5079 & 42.352 & 0.9444 & 0.7845 & 0.7768 & 0.7384 \\
		& SwinLSTM (ICCV’2023) & 20.19M & 69.87G & 4.3614 & 46.550 & 0.9333 & 0.7568 & 0.7505 & 0.7173 \\
		\midrule
		\multirow{7}{*}{\rotatebox{90}{\textbf{Recurrent-free}}}
		& SimVP-gSTA (CVPR’2022) & 46.77M & 16.53G & 7.5089 & 58.147 & 0.9037 & 0.6978 & 0.6897 & 0.6635 \\
		& SimVP-ViT (NeurIPS’2023) & 11.48M & 16.99G & 10.6166 & 75.952 & 0.8540 & 0.6271 & 0.6069 & 0.5667 \\
		& TAU (CVPR’2023) & 44.65M & 15.95G & 8.8433 & 62.247 & 0.8914 & 0.6744 & 0.6662 & 0.6428 \\
		& MMVP (ICCV’2023) & 14.82M & 93.55G & 9.3767 & 71.070 & 0.8688 & 0.6498 & 0.6347 & 0.5973 \\
		& WaST (AAAI’2024) & 28.18M & 21.90G & 6.0393 & 54.299 & 0.9156 & 0.7205 & 0.7062 & 0.6712 \\
		& Predformer-TSST (Arxiv’2025) & 25.30M & 16.47G & 4.1106 & 44.899 & 0.9391 & 0.7755 & 0.7697 & 0.7358 \\
		& Predformer-STTS (Arxiv’2025) & 25.30M & 16.47G & 7.3239 & 61.378 & 0.8987 & 0.6969 & 0.6834 & 0.6476 \\
		\midrule
		\multirow{5}{*}{\rotatebox{90}{\textbf{Ours}}}
		& \textbf{PredRNN + FB} & 25.01M & 0.121T & 3.2626 & 39.194 & 0.9496 & 0.7969 & 0.7922 & 0.7553 \\
		& \textbf{PredRNNpp + FB} & 39.76M & 0.177T & \textbf{2.7818} & \textbf{36.974} & \textbf{0.9562} & \textbf{0.8137} & \textbf{0.8080} & \textbf{0.7705} \\
		& \textbf{MIM + FB} & 38.92M & 0.186T & 2.9972 & \underline{37.009} & 0.9542 & 0.8076 & 0.8008 & 0.7606 \\
		& \textbf{MotionRNN + FB} & 28.09M & 0.134T & 3.2128 & 40.654 & 0.9504 & 0.7996 & 0.7940 & 0.7549 \\
		& \textbf{PredRNN-V2 + FB} & 25.03M & 0.122T & 3.1184 & 38.480 & 0.9509 & 0.8016 & 0.7974 & 0.7594 \\
		\bottomrule
	\end{tabular}
	\label{tab:shanghai2020}
\end{table*}

On the CIKM2017 dataset, we also evaluated the performance of various temporal prediction models. As shown in Table \ref{tab:cikm2017}, traditional recurrent models (such as PredRNN, PredRNN++, MIM, MotionRNN, etc.) achieved relatively strong results on the rainfall prediction task but still left room for further improvement in MSE, MAE, SSIM, and the CSI metrics at different thresholds. After integrating the proposed FlashBack (FB) module into these baseline models, overall performance improved to varying degrees. For example, PredRNN + FB reduced MSE from 28.81 to 27.15, lowered MAE from 159.40 to 155.79, increased SSIM to 0.7403, and consistently improved the CSI scores; MIM + FB decreased MSE from 27.21 to 26.62 while maintaining a high SSIM (0.7407) and stable CSI; MotionRNN + FB also achieved slight gains in MSE, MAE, and SSIM. Overall, although the CIKM2017 dataset has higher resolution and more complex variability compared to Shanghai2020, the FB module still demonstrated strong generalization capability, delivering consistent improvements across multiple metrics and validating its effectiveness and versatility in different rainfall prediction tasks.

On the SEVIR dataset (Table \ref{tab:sevir}), the precipitation echo sequences span longer periods and exhibit more complex weather processes, resulting in overall higher errors for the baseline models compared to the previous two datasets. Traditional recurrent methods such as PredRNN, PredRNN++, MIM, and MotionRNN show notable differences in MSE, MAE, and SSIM, and the CSI30/40/50 scores decrease significantly as the threshold increases, reflecting the difficulty of predicting heavy rainfall events. After integrating the proposed FlashBack (FB) module into these models, performance generally improved. For instance, PredRNN + FB reduced MSE from 53.49 to 49.68, MAE to 198.77, increased SSIM to 0.6999, and improved all CSI metrics; MotionRNN + FB also outperformed its original counterpart across multiple metrics; PredRNNpp + FB achieved lower MSE (52.45 vs. 57.69) and higher SSIM (0.6993 vs. 0.6930). These results indicate that FB can effectively enhance the spatiotemporal feature extraction capability of models in complex precipitation scenarios, improving the capture of heavy rainfall events and overall prediction quality.

\renewcommand{\arraystretch}{1.2} 
\begin{table*}[htbp]
	\centering
	\caption{Quantitative results on CIKM2017 datasets.}
	\begin{tabular}{c|l|cccccccc}
		\specialrule{1.5pt}{1pt}{1pt}
		\textbf{Type} & \textbf{Model} & \textbf{Params} & \textbf{FLOPS} & \textbf{MSE} $\downarrow$ & \textbf{MAE} $\downarrow$ & \textbf{SSIM} $\uparrow$ & \textbf{CSI30} $\uparrow$ & \textbf{CSI40} $\uparrow$ & \textbf{CSI50} $\uparrow$ \\
		\specialrule{1pt}{1pt}{1pt} 
		\midrule
		\multirow{9}{*}{\rotatebox{90}{\textbf{Recurrent-based}}} 
		& ConvLSTM (NeurIPS’2015) & 15.1M & 41.84G & 28.8162 & 161.242 & 0.7359 & 0.7757 & 0.6590 & 0.5367 \\
		& PredRNN (NeurIPS’2017) & 23.8M & 85.43G & 28.8126 & 159.404 & 0.7303 & 0.7803 & 0.6690 & 0.5471 \\
		& PredRNNpp (ICML’2018) & 38.6M & 0.127T & 28.0987 & 159.258 & 0.7362 & 0.7775 & 0.6684 & 0.5533 \\
		& MIM (CVPR’2019) & 38.04M & 0.131T & 27.2072 & 154.269 & 0.7388 & 0.7828 & 0.6725 & \underline{0.5587} \\
		& PyDNet (CVPR’2020) & 3.1M & 11.29G & 28.9757 & 161.143 & 0.7393 & 0.7809 & 0.6654 & 0.5489 \\
		& MotionRNN (CVPR’2021) & 26.9M & 94.80G & 27.2091 & 155.827 & 0.7406 & \underline{0.7867} & \underline{0.6762} & 0.5510 \\
		& MAU (NeurIPS’2021) & 4.48M & 13.11G & 30.9094 & 167.361 & 0.7234 & 0.7731 & 0.6584 & 0.5459 \\
		& PredRNN-V2 (TPAMI’2022) & 23.9M & 85.91G & 28.0085 & 161.508 & 0.7374 & 0.7857 & 0.6640 & 0.5362 \\
		& SwinLSTM (ICCV’2023) & 20.19M & 51.49G & 27.9612 & 158.899 & 0.7405 & 0.7826 & 0.6652 & 0.5532 \\
		\midrule
		\multirow{7}{*}{\rotatebox{90}{\textbf{Recurrent-free}}}
		& SimVP-gSTA (CVPR’2022) & 4.82M & 1.238G & 31.3121 & 166.261 & 0.7195 & 0.7731 & 0.6594 & 0.5416 \\
		& SimVP-ViT (NeurIPS’2023) & 39.6M & 12.87G & 27.8803 & 157.478 & 0.7358 & 0.7835 & 0.6669 & 0.5483 \\
		& TAU (CVPR’2023) & 38.4M & 12.01G & 30.5827 & 161.948 & 0.7277 & 0.7779 & 0.6601 & 0.5370 \\
		& MMVP (ICCV’2023) & 12.98M & 69.45G & 30.1724 & 165.321 & 0.7285 & 0.7826 & 0.6730 & \textbf{0.5602} \\
		& WaST (AAAI’2024) & 28.18M & 14.56G & 30.3074& 165.804 & 0.7309 & 0.7773 & 0.6534 & 0.5174 \\
		& Predformer-TSST (Arxiv’2025) & 25.30M & 8.23G & 26.7383 & 163.146 & 0.7316 & 0.7872 & 0.6738 & 0.5522 \\
		& Predformer-STTS (Arxiv’2025) & 25.30M & 8.23G & 27.3831 & 168.386 & 0.7143 & 0.7807 & 0.6638 & 0.5473 \\
		\midrule
		\multirow{5}{*}{\rotatebox{90}{\textbf{Ours}}}
		& \textbf{PredRNN + FB} & 25.02M & 89.36G & 27.1508 & 155.792 & 0.7403 & 0.7857 & \textbf{0.6767} & 0.5566 \\
		& \textbf{PredRNNpp + FB} & 39.76M & 0.130T & 27.8036 & 157.915 & 0.7393 & 0.7825 & 0.6724 & 0.5553 \\
		& \textbf{MIM + FB} & 39.22M & 0.135T & \textbf{26.6192} & \textbf{153.506} & 0.7407 & 0.7815 & 0.6720 & 0.5534 \\
		& \textbf{MotionRNN + FB} & 28.09M & 98.73G & \underline{26.6759} & \underline{154.056} & \textbf{0.7438} & 0.7838 & 0.6745 & 0.5570 \\
		& \textbf{PredRNN-V2 + FB} & 25.03M & 89.83G & 27.2468 & 158.862 & \underline{0.7413} & \textbf{0.7873} & 0.6707 & 0.5467 \\
		\bottomrule
	\end{tabular}
	\label{tab:cikm2017}
\end{table*}

\renewcommand{\arraystretch}{1.2} 
\begin{table*}[htbp]
	\centering
	\caption{Quantitative results on SEVIR datasets. *(Due to computational resource limitations, we only trained SwinLSTM-B here.)}
	\begin{tabular}{c|l|cccccccc}
		\specialrule{1.5pt}{1pt}{1pt}
		\textbf{Type} & \textbf{Model} & \textbf{Params} & \textbf{FLOPS} & \textbf{MSE} $\downarrow$ & \textbf{MAE} $\downarrow$ & \textbf{SSIM} $\uparrow$ & \textbf{CSI30} $\uparrow$ & \textbf{CSI40} $\uparrow$ & \textbf{CSI50} $\uparrow$ \\
		\specialrule{1pt}{1pt}{1pt} 
		\midrule
		\multirow{9}{*}{\rotatebox{90}{\textbf{Recurrent-based}}} 
		& ConvLSTM (NeurIPS’2015) & 15.08M & 71.735G & \textbf{52.3914} & 203.373 & 0.6938 & 0.4469 & 0.3842 & 0.3182 \\
		& PredRNN (NeurIPS’2017) & 23.84M & 0.146T & 53.4924 & 208.914 & 0.6826 & 0.4188 & 0.3656 & 0.3116 \\
		& PredRNNpp (ICML’2018) & 38.58M & 0.217T & 57.6933 & 208.626 & 0.6930 & 0.4372 & 0.3823 & \underline{0.3265} \\
		& MIM (CVPR’2019) & 38.34M & 0.229T & 61.5472 & 219.087 & 0.6753 & 0.4094 & 0.3559 & 0.3003 \\
		& PyDNet (CVPR’2020) & 3.09M & 19.361G & 57.1866 & 225.926 & 0.6365 & 0.4139 & 0.3491 & 0.2806 \\
		& MotionRNN (CVPR’2021) & 26.91M & 0.163T & 60.6225 & 214.475 & 0.6824 & 0.4293 & 0.3764 & 0.3212 \\
		& MAU (NeurIPS’2021) & 4.48M & 22.47G & 64.6087 & 222.400 & 0.6801 & 0.4152 & 0.3629 & 0.3089 \\
		& PredRNN-V2 (TPAMI’2022) & 23.85M & 0.147T & 56.1049 & 232.900 & 0.6255 & 0.3940 & 0.3290 & 0.2592 \\
		& SwinLSTM* (ICCV’2023) & 0.7M & 16.119G & 55.5566 & 224.417 & 0.6481 & 0.4062 & 0.3369 & 0.2773 \\
		\midrule
		\multirow{7}{*}{\rotatebox{90}{\textbf{Recurrent-free}}}
		& SimVP-gSTA (CVPR’2022) & 46.77M & 16.53G & 60.6290 & 206.135 & 0.6938 & 0.4299 & 0.3742 & 0.3163 \\
		& SimVP-ViT (NeurIPS’2023) & 11.48M & 16.99G & 59.9962 & 225.128 & 0.6475 & 0.4112 & 0.3445 & 0.2756 \\
		& TAU (CVPR’2023) & 44.65M & 15.96G & 61.2723 & 208.601 & 0.6924 & 0.4315 & 0.3748 & 0.3168 \\
		& MMVP (ICCV’2023) & 16.49M & 0.12T & 62.9421 & 256.710 & 0.5860 & 0.3483 & 0.2815 & 0.2084 \\
		& WaST (AAAI’2024) & 28.18M & 21.90G & 54.3905 & \textbf{196.556} & \textbf{0.7134} & \underline{0.4534} & 0.3884 & 0.3233 \\
		& Predformer-TSST (Arxiv’2025) & 25.30M & 16.478G & 55.8347 & 217.876 & 0.6615 & 0.4071 & 0.3451 & 0.2839 \\ 
		& Predformer-STTS (Arxiv’2025) & 25.30M & 16.478G & 57.9686 & 224.831 & 0.6280 & 0.3681 & 0.2969 & 0.2279 \\
		\midrule
		\multirow{5}{*}{\rotatebox{90}{\textbf{Ours}}}
		& \textbf{PredRNN + FB} & 25.02M & 0.153T & 49.6812 & 198.773 & 0.6999 & \textbf{0.4538} & \underline{0.3895} & 0.3251 \\
		& \textbf{PredRNNpp + FB} & 39.76M & 0.224T & \underline{52.4572} & 201.005 & \underline{0.6993} & 0.4499 & 0.3863 & 0.3230 \\
		& \textbf{MIM + FB} & 38.93M & 0.232T & 58.1526 & 215.574 & 0.6758 & 0.4185 & 0.3616 & 0.3040 \\
		& \textbf{MotionRNN + FB} & 28.092M & 0.169T & 52.5389 & \underline{199.990} & 0.6996 & 0.4521 & \textbf{0.3902} & \textbf{0.3271} \\
		& \textbf{PredRNN-V2 + FB} & 25.03M & 0.154T & 56.1824 & 230.289 & 0.6300 & 0.3983 & 0.3318 & 0.2611 \\
		\bottomrule
	\end{tabular}
	\label{tab:sevir}
\end{table*}

Across the three precipitation datasets, integrating the FlashBack (FB) module consistently enhances the performance of baseline models in precipitation nowcasting tasks. On the Shanghai2020 dataset, FB improves MSE, MAE, SSIM, and CSI metrics, demonstrating its effectiveness in capturing short- and medium-term rainfall dynamics. On the CIKM2017 dataset, despite higher spatial resolution and more complex temporal variations, FB achieves consistent improvements across key metrics, highlighting its strong generalization capability. In the more challenging SEVIR dataset, where sequences are longer and weather processes are more complex, the FB module substantially reduces errors and enhances structural similarity, particularly in the prediction of heavy rainfall events, as reflected in the improved CSI scores. Overall, the FB module robustly strengthens the spatiotemporal feature modeling ability of the models, improving quantitative prediction accuracy across different rainfall datasets while enhancing the capability to capture extreme weather events.

\subsection{Qualitative Evaluation}

\begin{figure*}[htbp]
	\centerline{\includegraphics[width=18cm]{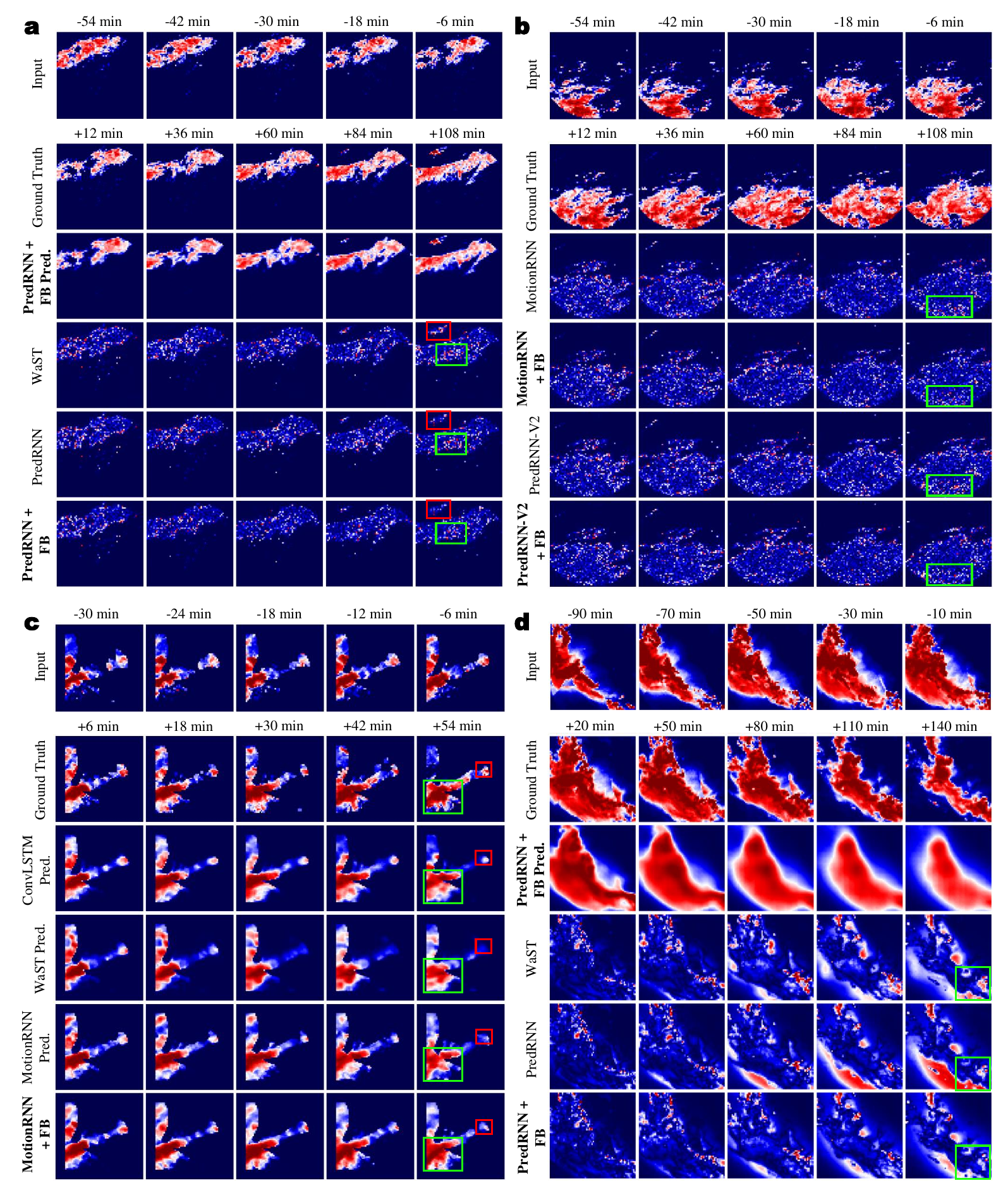}}
	\caption{Qualitative Visualization Results on Shanghai2020, CIKM2017, and SEVIR Datasets. In the figure, the labeled model name “pred.” represents the model’s actual prediction results, while the labeled model name with error annotations indicates the absolute error between the model’s prediction and the ground truth. (a) and (b) correspond to the Shanghai2020 dataset, (c) corresponds to the CIKM2017 dataset, and (d) corresponds to the SEVIR dataset.
	}
	\label{fig:three}
\end{figure*}

Figure \ref{fig:three}-a shows a precipitation case from the Shanghai2020 dataset. We compare the absolute prediction errors (AE) of WaST, PredRNN, and PredRNN+FB to analyze their forecasting performance. Two main regions with concentrated errors are highlighted (green and red boxes). In the green box, WaST and PredRNN exhibit obvious false alarms of precipitation, whereas PredRNN+FB successfully suppresses such noise and produces results closer to the ground truth. In the red box, where rainfall gradually emerges, WaST suffers from severe missed detections, while PredRNN+FB effectively captures the evolution of this dynamic event. This demonstrates that the FB module can significantly enhance the model’s ability to capture localized heavy-rainfall dynamics.

Figure \ref{fig:three}-b presents another case from Shanghai2020, where we compare the prediction errors of PredRNN and MotionRNN before and after integrating FB. As shown in the green-boxed regions with concentrated errors, both baseline models achieve markedly reduced prediction bias after adding FB, and their spatial details more closely resemble the true rainfall field. This indicates that FB provides stronger feature extraction and dynamic variation capture capabilities when forecasting rainfall events with complex spatial distributions and evolving patterns.

Figure \ref{fig:three}-c displays a forecasting case from the CIKM2017 dataset. Compared with Shanghai2020, CIKM2017 is more challenging due to less input information. In the green box, ConvLSTM, WaST, and MotionRNN all substantially underestimate rainfall intensity; in the red box, WaST and MotionRNN fail to report rainfall in that region. By contrast, MotionRNN+FB achieves better forecasts of both rainfall intensity and location in these two areas, showing that the FB module can effectively enhance spatiotemporal feature learning and improve the detection accuracy and spatial localization of heavy rainfall events under limited input and more complex forecasting conditions.

Figure \ref{fig:three}-d shows a forecasting case from the SEVIR dataset. Compared with the previous two datasets, SEVIR features longer temporal spans and more complex weather evolutions, placing higher demands on long-sequence precipitation forecasting. We compare the absolute prediction errors of WaST, PredRNN, and PredRNN+FB. The results show that the original PredRNN produces large errors in long-sequence prediction, whereas integrating FB significantly reduces the bias. Particularly in the green-boxed focus area, PredRNN+FB outperforms both the baseline model and WaST. This demonstrates that the FB module can effectively strengthen spatiotemporal feature extraction and memory in long-sequence, complex precipitation scenarios, markedly improving the capture accuracy of key rainfall regions and the overall forecast quality.

Overall, the case studies in Figure \ref{fig:three} across Shanghai2020, CIKM2017, and SEVIR clearly demonstrate the consistent effectiveness of the proposed FB module. By integrating FB into various baseline models, prediction errors are substantially reduced, false alarms are suppressed, and dynamic rainfall processes are more accurately captured. These improvements hold across datasets of increasing complexity and longer temporal spans, indicating that FB not only enhances short-term and mid-scale rainfall forecasts but also significantly boosts long-sequence prediction capability. This highlights FB as a robust, generalizable enhancement for spatiotemporal modeling in precipitation forecasting tasks.

\begin{figure*}[htbp]
	\centerline{\includegraphics[width=18cm]{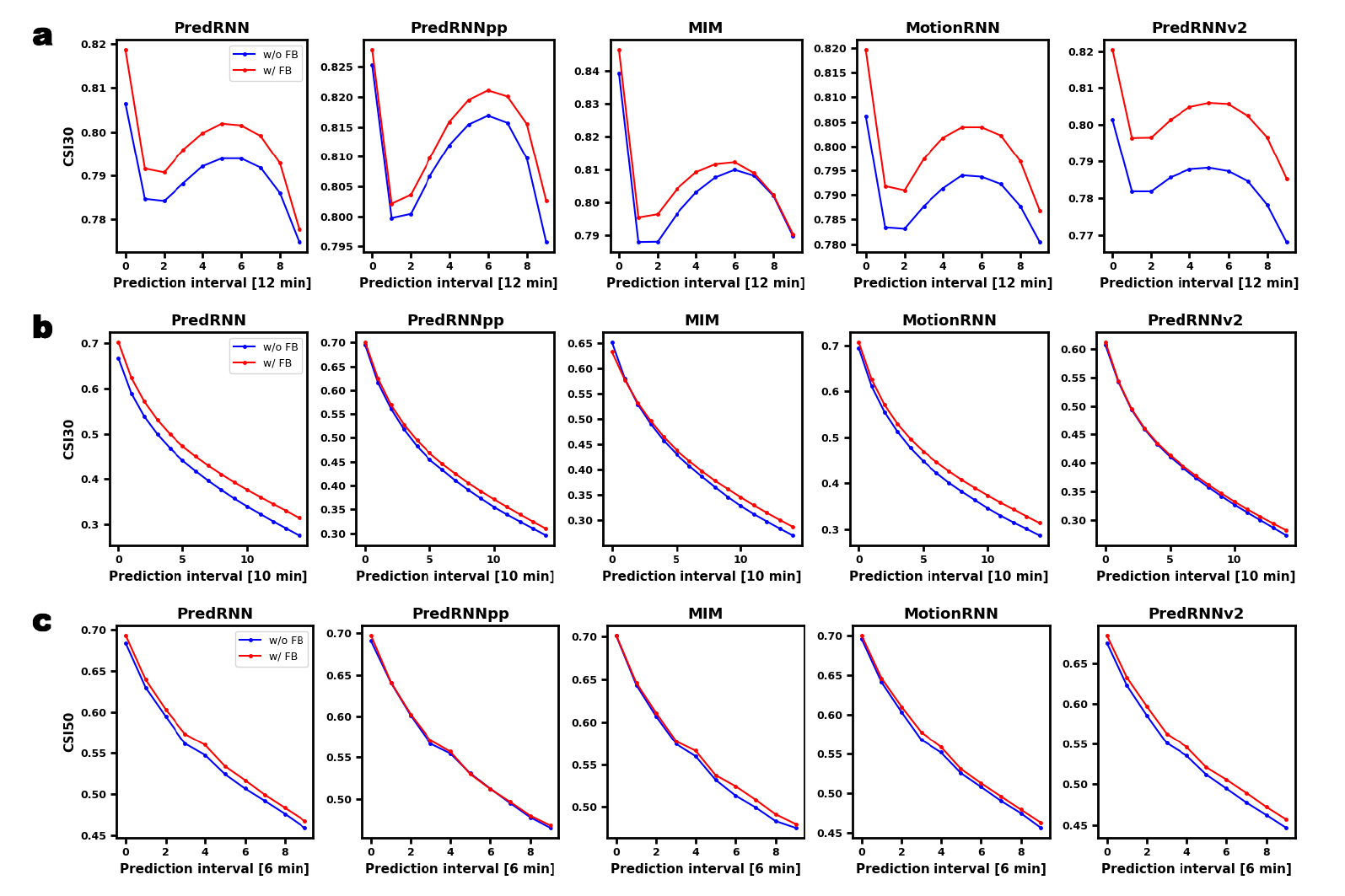}}
	\caption{Changes in CSI metrics of baseline models before and after integrating FB. (a), (b), and (c) correspond to the Shanghai2020, SEVIR and CIKM2017 datasets, respectively.
	}
	\label{fig:four}
\end{figure*}

To further verify the improvement brought by the FlashBack module to Recurrent-based models in capturing extreme precipitation events, we compared the changes in CSI metrics of the baseline models (PredRNN, PredRNNpp, MIM, MotionRNN, PredRNN-V2) before and after adding the FB module on the Shanghai2020, CIKM2017, and SEVIR datasets (Figure \ref{fig:four}). It is evident that the CSI of all models improved after integrating the FB module, with more pronounced gains observed at higher thresholds (e.g., CSI40, CSI50). This indicates that the FB module effectively enhances the ability of Recurrent-based models to capture intense rainfall events and further validates its generalizability and performance improvement across different precipitation datasets.

\subsection{Ablation Study}

To further investigate the impact of the length-based hyperparameter $K_{\text{max}}$ in the FlashBack (FB) module on prediction performance, we conducted an ablation study on PredRNN using the Shanghai2020 dataset, as shown in the table. It can be observed that as $K_{\text{max}}$ increases from 1 to 10, the model’s MSE and MAE steadily decrease, while SSIM and CSI at all thresholds improve, indicating that a larger $K_{\text{max}}$ helps the FB module capture longer-term temporal dynamics and thus enhances precipitation prediction accuracy. However, when $K_{\text{max}}$ increases from 10 to 15, the improvements plateau or slightly decline, suggesting that an excessively large $K_{\text{max}}$ does not provide significant benefits and may introduce slight redundancy due to increased model complexity. Therefore, in this experiment, $K_{\text{max}} = 10$ represents a reasonable trade-off between performance and computational cost.

\begin{table}[htbp]
	\small
	\centering
	\caption{Ablation Study of $K_{\text{max}}$ for PredRNN + FB on the Shanghai2020 Dataset}
	\resizebox{0.95\linewidth}{!}{%
		\begin{tabular}{ccccccc}
			\toprule
			$\boldsymbol{K}_{\text{max}}$ & \textbf{MSE} $\downarrow$ & \textbf{MAE} $\downarrow$ & \textbf{SSIM}$\uparrow$ & \textbf{CSI30}$\uparrow$ & \textbf{CSI40} $\uparrow$ & \textbf{CSI50} $\uparrow$ \\
			\midrule
			1                & 3.4141 & 39.849 & 0.9475 & 0.7918 & 0.7873 & 0.7506 \\
			3                & 3.3391 & 39.636 & 0.9487 & 0.7950 & 0.7901 & 0.7534 \\
			5                & 3.2919 & 39.400 & 0.9493 & 0.7962 & 0.7913 & 0.7544 \\
			10               & 3.2626 & 39.194 & 0.9496 & 0.7969 & 0.7922 & 0.7553 \\
			15               & 3.2840 & 39.332 & 0.9495 & 0.7961 & 0.7914 & 0.7545 \\
			\bottomrule
		\end{tabular}%
	}
	\label{tab:predrnn_kmax_ablation}
\end{table}

\begin{figure}[htbp]
	\centerline{\includegraphics[width=9cm]{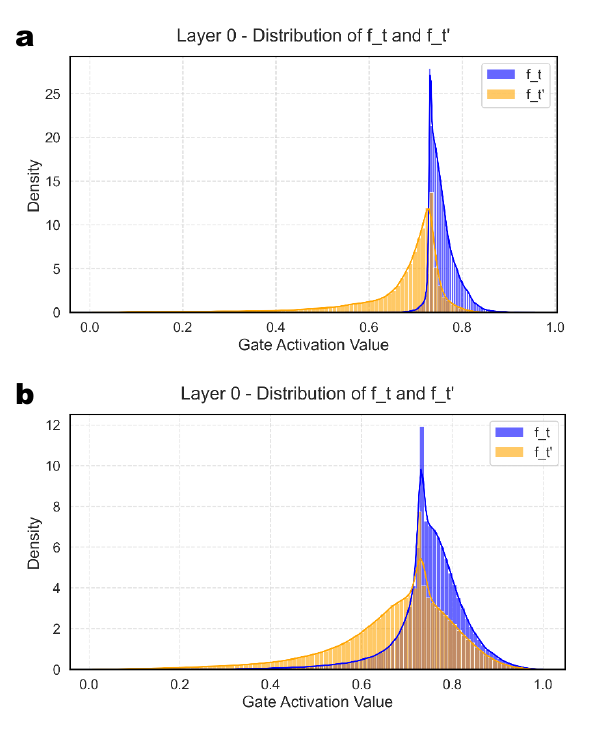}}
	\caption{Feature distribution comparison on CIKM2017. (a) $f_t$ and $f_t'$ in the first layer of PredRNN. (b) $f_t$ and $f_t'$ in the first layer of PredRNN+FB.
	}
	\label{fig:five}
\end{figure}

To further investigate the effect of the FlashBack (FB) module on the internal feature representations of recurrent models, we analyzed the distributions of the temporal feature $f_t$ and the memory-fused feature $f_t'$ of PredRNN on the CIKM2017 dataset (Figure \ref{fig:five}).  The results indicate that the FB module can significantly enhance PredRNN's capability in modeling sequential rainfall events. By comparing the distributions of $f_t$ and the historical memory feature $f_t'$ in the first hidden layer of PredRNN and PredRNN+FB, we observed that the mean of $f_t$ in the original PredRNN is approximately 0.73, and the mean of $f_t'$ is about 0.62. After incorporating FB, the mean of $f_t$ slightly decreases to 0.69, while the mean of $f_t'$ increases to 0.63. This suggests that the FB module, through dynamic routing and feature fusion, transfers part of the information from the current frame feature $f_t$ to the historical memory $f_t'$, achieving a more balanced distribution of current and historical information. Although the mean of $f_t$ slightly decreases, this does not indicate a performance drop; rather, it reflects a "denoising" effect on the current frame features while improving temporal consistency and capturing dynamic rainfall events. Overall, the FB module can effectively enhance the spatiotemporal feature modeling and prediction accuracy of Recurrent-based models for extreme rainfall events without imposing significant additional computational overhead.

\section{Conclusion}
In this work, we propose FlashBack Memory(FB), a memory-enhanced module designed to improve the spatiotemporal modeling capability of recurrent-based precipitation prediction models. FB leverages dynamic routing and historical feature fusion to selectively retrieve and integrate relevant past information, enabling the model to better capture complex rainfall dynamics and extreme precipitation events. Extensive experiments on three benchmark datasets—Shanghai2020, CIKM2017, and SEVIR—demonstrate that integrating FB consistently improves the performance of baseline models such as PredRNN, PredRNNpp, MIM, MotionRNN, and PredRNN-V2 across multiple metrics including MSE, MAE, SSIM, and CSI at different thresholds. Visualization analyses further reveal that FB enhances the temporal consistency of predictions, reduces false alarms and missed events, and enables more accurate spatial localization of rainfall patterns. Ablation studies on the memory length hyperparameter confirm the effectiveness of FB in balancing current-frame and historical memory features. Overall, our approach provides a general, lightweight, and effective mechanism to boost the predictive performance of recurrent-based models for both short-term and long-term rainfall forecasting, particularly in challenging high-resolution and complex precipitation scenarios.

\end{document}